\documentclass[letterpaper, 10 pt, conference]{ieeeconf}

\IEEEoverridecommandlockouts

\usepackage{url}
\usepackage{balance}

\usepackage{amsmath}
\usepackage{amssymb}
\usepackage{amsfonts}
\usepackage{enumerate}
\usepackage{tabulary}
\usepackage{multirow}
\usepackage{cite}
\usepackage{lipsum}  
\usepackage{graphicx}
\usepackage{epstopdf}
\usepackage{filecontents}
\usepackage[misc]{ifsym}
\usepackage{color}
\usepackage{xcolor}
\usepackage{xxcolor}
\usepackage{pgf}
\usepackage{pgfplots}
\usepackage{tikz}
\usepackage{float}
\usepackage{ifthen}
\usepackage{forloop}

\usepgfplotslibrary{groupplots,fillbetween}
\usetikzlibrary{arrows,shadows,petri,automata,shapes.geometric}
\pgfplotsset{compat=newest}

\usepackage{listings}
\usepackage{lipsum}
\usepackage{booktabs}
\usepackage[lined,boxed,commentsnumbered,linesnumbered]{algorithm2e}
\usetikzlibrary{positioning,fit,calc}
\usetikzlibrary{arrows.meta}
\usetikzlibrary{spy}
\usepackage{circuitikz}
\SetAlCapSkip{0.5em}
\definecolor{codegreen}{rgb}{0,0.6,0}
\definecolor{codepurple}{rgb}{0.58,0,0.82}
\definecolor{backcolour}{rgb}{0.95,0.95,0.92}
\lstdefinestyle{buzz}{
    backgroundcolor=\color{black!5},   
    commentstyle=\color{codegreen},
    keywordstyle=\color{blue},
    numberstyle=\tiny\color{black!30},
    stringstyle=\color{codepurple},
    basicstyle=\footnotesize\ttfamily,
    breakatwhitespace=false,         
    breaklines=true,                 
    captionpos=b,                    
    keepspaces=true,                 
    numbers=left,                    
    numbersep=5pt,                  
    showspaces=false,                
    showstringspaces=false,
    showtabs=false,                  
    tabsize=2,
}
\title{\LARGE \bf
Decentralized Connectivity Control in Quadcopters: \\
a Field Study of Communication Performance
}

\author{Jacopo Panerati$^{\dag}$, Benjamin Ramtoula$^{\dag}$, David St-Onge$^{\dag}$, Yanjun Cao$^{\dag}$, Marcel Kaufmann$^{\dag}$,\\ Aidan Cowley$^{*}$, Lorenzo Sabattini$^{\ddag}$, and Giovanni Beltrame$^{\dag}$
\thanks{$^{\dag}$Polytechnique Montr\'eal, 
              Department of Software and Computer Engineering, Montreal, Canada,
        e-mail:
        {\tt \{name.lastname\}@polymtl.ca}}%
\thanks{$^{*}$European Space Agency, European Astronaut Centre, K\"oln, Germany, 
        e-mail:
        {\tt adian.cowley@esa.int}}%
\thanks{$^{\ddag}$Universit\`a degli Studi di Modena e Reggio Emilia, Department of Sciences and Methods for Engineering, Reggio Emilia, Italy, %
e-mail: {\tt lorenzo.sabattini@unimore.it}}%
}

\begin{document}

\maketitle
\thispagestyle{empty}
\pagestyle{empty}

\begin{abstract}
  Redundancy and parallelism make decentralized multi-robot systems
  appealing solutions for the exploration of extreme environments.  However,
  effective cooperation often requires team-wide connectivity and a carefully
  designed communication strategy.  Several recently proposed decentralized
  connectivity maintenance approaches exploit elegant algebraic results drawn
  from spectral graph theory.  Yet, these proposals are rarely taken beyond
  simulations or laboratory implementations.  In this work, we present two
  major contributions: (i) we describe the full-stack implementation---from
  hardware to software---of a decentralized control law for robust
  connectivity maintenance; and (ii) we assess, in the field, our setup's ability to correctly exchange all the
  necessary information required to maintain connectivity in a team of quadcopters.
\end{abstract}

\section{Introduction}
\label{sec:intro}

Multi-robot systems can be used to tackle complex problems that benefit from
physical parallelism and the inherent fault-tolerance provided by
redundancy---surveillance, disaster recovery, and planetary exploration being
a few notable examples.  Decentralized control strategies further
improve the reliability of these systems by partially relaxing communication
bandwidth requirements and eliminating the risks posed by single points of
failure.  Swarm robotics~\cite{sahin2008} is the branch of
robotics focusing on decentralized many-robot systems.  Complementarily, swarm
intelligence research aims at overcoming the limited capabilities of swarms'
individual agents through the design of intelligent coordination.

For many multi-robot applications, an essential requirement for effective
cooperation is the enforcement of global connectivity.  That is, the ability
for every robot to find a communication path to any other robot in the
team.  When only limited-range communication is available, global connectivity
can require intermediate robots to also act as relays.  Assessing and
controlling the global connectivity of a communication graph (where robots are
nodes and radios create links) in a decentralized fashion is not
trivial~\cite{bertrand2013a}.  Several recent
approaches~\cite{zavlanos2011,robuffogiordano2013,solana2017} exploit the
spectral graph theory result stating that the second smallest eigenvalue of
the Laplacian matrix $L$ of the communication graph (often referred to as
$\lambda_2$, $\lambda$, or algebraic connectivity), is non-zero \emph{if and
  only if} the underlying communication graph is connected~\cite{fiedler1973}.
These proposals, however, are typically limited to
simulations~\cite{zavlanos2011} or laboratory
experiments~\cite{robuffogiordano2013}.

In this work, we provide two contributions to the research on decentralized
assessment and control of algebraic connectivity (and, in general, multi-robot
connectivity maintenance).  First, we present how to implement
a decentralized, robust, connectivity control law~\cite{ghedini2017b} in a
team of quadcopters---from the computing and communication hardware level, to the robotic
middleware and control software.  Second, we report field experiments
conducted by flying three quadcopters implementing this hardware and software
stack.  Our results show that, despite the presence of an expected reality
gap, our setup can successfully exchange the information required by the
decentralized control law.

The rest of the article is organized as follows: Section~\ref{sec:related}
briefly reviews the state-of-the-art in decentralized connectivity control;
Section~\ref{sec:law} presents the specific control law under scrutiny in this
work; then, Section~\ref{sec:field} describes its practical implementation in
a team of quadcopters for field testing.  Finally, performance results are
given in Section~\ref{sec:results} and Section~\ref{sec:conclusions} concludes
the article.

\section{Related Work}
\label{sec:related}

Algebraic connectivity is a well-established graph theory concept.  Miroslav
Fiedler wrote about the properties of the second smallest eigenvalue
$\lambda_2$---also called Fiedler eigenvalue---of the unweighted Laplacian
matrix of a graph in a seminal paper~\cite{fiedler1973} where he derives from the Perron--Frobenius theorem, that $\lambda_2$ ``is zero
if and only if the graph is not connected''.  On the other hand, more recent
research has proposed approaches for its computation in a decentralized
fashion in ad-hoc networks.  The work of Sahai \emph{et
  al.}~\cite{sahai2012a}, for example, exploits wave propagation and fast
Fourier transforms while Bertrand and Moonen~\cite{bertrand2013a} propose a
method based on the power iteration algorithm.

As multi-robot systems research proliferated over the last decade, many
suggested to include algebraic connectivity in control laws aimed at
preserving the global connectivity~\cite{panerati2018icra} of robotic teams.  Ji and
Egerstedt~\cite{ji2007} proposed---and evaluated in simulation---multiple
feedback control laws ensuring connectivity for the rendezvous and formation
control problems based on the weighted Laplacian matrix.  Zavlanos \emph{et
  al.}~\cite{zavlanos2011} presented centralized and distributed approaches to
algebraic connectivity maximization, adding flocking to the two control
problems in~\cite{ji2007} and also providing simulation results.  Robuffo
Giordano \emph{et al.}~\cite{robuffogiordano2013} introduced a decentralized
control law based on a potential function of algebraic connectivity. Their
work was tested with four quadrotors in a laboratory setting (using Wi-Fi for
communication and a commercial mo-cap solution for localization).  Even so,
the authors observed discrepancies ``due to the presence of noise and small
communication delays, and in general to all of those non-idealities and
disturbances affecting real conditions''~\cite{robuffogiordano2013}.
Sabattini \emph{et al.}~\cite{Sabattini:2013} evaluated their decentralized
connectivity maintenance control law using four E-Puck robots.  Solana
\emph{et al.}~\cite{solana2017} further advanced the research in generalized
connectivity control based on $\lambda_2$ accounting for path planning in
cluttered environments. Experiments with quadrotors were carried out in
simulation.

When aiming at field deployment in extreme areas (such as caves, planetary
surfaces, and regions hit by natural disasters), however, one has to make sure
that a control law is not only correct in nominal situations but its
performance is also robust against hardware and communication failures. When
it comes to connectivity, this means that approaches only controlling the
Fielder eigenvalue might be unsuccessful as they can be blind to certain
pathological configurations with highly vulnerable nodes.  A combined control
law---to simultaneously improve algebraic connectivity and robustness of a
network---was proposed and evaluated in simulation by Ghedini \emph{et al.}~\cite{ghedini2017b}.
We brought this approach to a real-world implementation using eight K-Team
Khepera IV robots and tested against faulty communication---albeit only
through emulation---in~\cite{panerati2018}.  Finer tuning of its
hyper-parameterization and coverage approach were discussed
in~\cite{minelli2019} and~\cite{siligardi2019}, respectively.  The work in
this article advances
the state-of-the-art and our own previous
by investigating the challenges of transferring these approaches
beyond the reality gap and into the domain of field robotics.

\section{Control Law}
\label{sec:law}

We consider the control law proposed in~\cite{ghedini2017b}. This law is
intended to both (i) preserve connectivity and (ii) strengthen the
robustness of the communication topology against the failure of individual
robots.  This control law can be implemented in a fully decentralized fashion
under the relatively loose and---in swarm robotics---common assumption of
exploiting the situated communication model~\cite{stoy2001a}.  This means
robots possess range and bearing information about their 1-hop neighbors
(see Figure~\ref{fig:theory}).  {Considering robots modeled as $m$-dimensional single integrators\footnote{{Even though this represents a very simple model, it is worth remarking that, by endowing a robot with a sufficiently good Cartesian trajectory tracking controller, the single integrator model can be exploited to represent the kinematic behavior of several types of mobile robots, like wheeled mobile robots~\cite{soukieh2009obstacle}, and
quadrotors~\cite{lee2013semiautonomous}.}}, and defining $p_i\in\mathbb{R}^m$ as the position of the $i$-th robot, the control law is defined as the %
linear
combination of connectivity, robustness, and coverage contributions
which, for robot $i$, can be written as:
\begin{equation}
	{\dot{p}_i=u_i=\sigma u_i^c + \psi u_i^r + \zeta u_i^{LJ}}
	\label{eq:law}
\end{equation}
The computation of {$u^c,u^r,u^{LJ}\in\mathbb{R}^m$} is detailed in the following subsections. 
Offline and online schemes for the selection of hyper-parameters {$\sigma,\psi,\zeta\in\mathbb{R}$} were presented in~\cite{panerati2018,minelli2019} and not further discussed here.

\begin{figure}%
	 \scalebox{.7}{
	\includegraphics[] {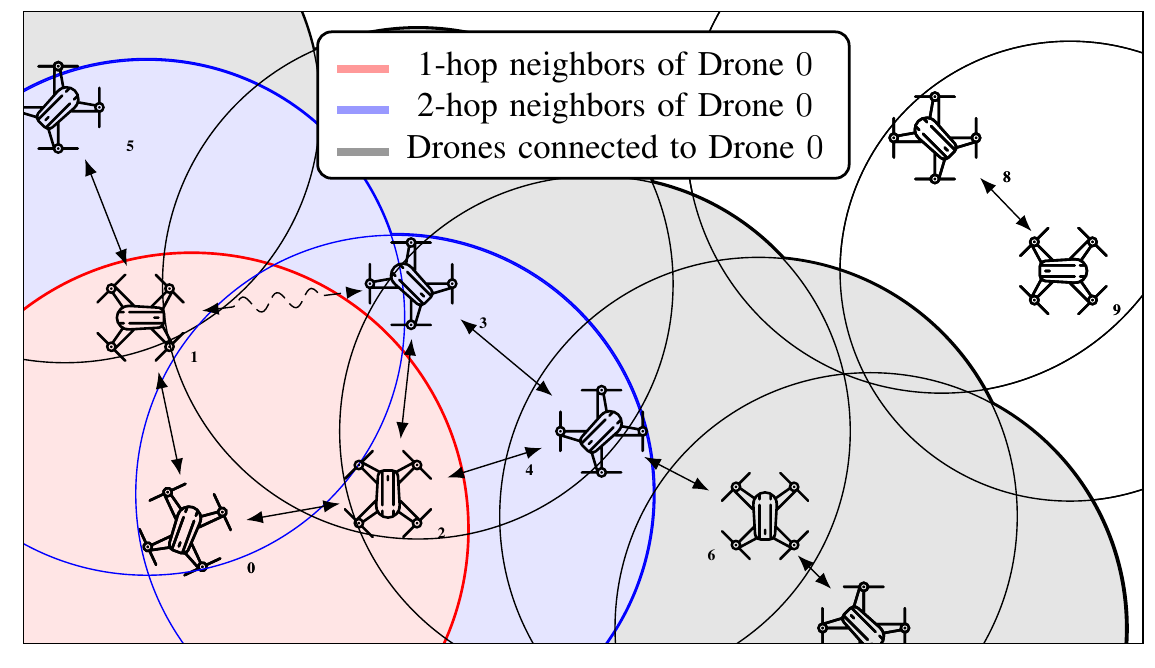}
	}
    
\caption{
In a multi-robot system with limited-range communication capabilities, we define as direct (or 1-hop) neighbors of a robot those robots that are within such range. We can then iteratively apply this notion to define 2-hop neighborhoods.
}
\label{fig:theory} 
\vspace{-1em}
\end{figure}

\subsection{Connectivity Maintenance Contribution}
\label{sec:connectivity}

The first component on the right side of \eqref{eq:law}, $u_i^c$, is
the one intended to maintain global connectivity, i.e., to prevent splits in
the communication graph of the multi-robot system.  Indeed, this is done
through the control of $\lambda_2$.  Algebraic connectivity is positive only
when the graph is connected and also upper bounds the sparsest cut in the
network.  Decentralized computation of $\lambda_2$ in ad-hoc networks was
demonstrated, among others, by~\cite{bertrand2013a} and~\cite{dilorenzo2014a}.
Both of these approaches rely on the power iteration (PI) algorithm: they
compute the largest eigenvalue (and associated eigenvector $\mathbf{x}$) of a
matrix $M$ using the update rule:
\begin{equation}
	\mathbf{x}^{l+1} = M  \mathbf{x}^{l}
	\label{eq:pi1} 
\end{equation}

Over communication graphs, the update in \eqref{eq:pi1} can be
computed in a decentralized fashion for any shift operator (i.e., any matrix
with the same sparsity pattern of the graph).  The adjacency $A$ and Laplacian
$L$ matrices are two such operators. For $L$ the
decentralized update rule becomes 
$$x^{l+1}_k = L_{kk} \cdot x^l_k + \sum_{j|j\ne k \land L_{kj}\ne0} L_{jk}
\cdot x^{l}_j$$
where $x^l_k$ is the $k$-th robot's estimate of the $k$-th entry of the eigenvector $\mathbf{x}$, at the $l$-th iteration, and $L_{kj}$ is the element $(k,j)$ of the Laplacian matrix $L$.

Bertrand and Moonen~\cite{bertrand2013a} showed how to derive a matrix $M$
from $L$ so that \eqref{eq:pi1} leads to $\lambda_2$.  Then, using an
energy function {$V(\lambda_2)$ that is non-negative, non-increasing with respect to $\lambda_2$, and that goes to infinity for $\lambda_2$ approaching zero (such as the one proposed in~\cite{Sabattini:2013})} , one can
compute the connectivity contribution to \eqref{eq:pi1} {as follows
\begin{equation}
	u_i^c = -\dfrac{\partial V\left(\lambda_2\right)}{\partial p_i}=-\dfrac{\partial V\left(\lambda_2\right)}{\partial \lambda_2}\dfrac{\partial \lambda_2}{\partial p_i}
	\label{eq:conn}
\end{equation}
}
The main caveat is that, as observed in~\cite{bertrand2013a}, a PI approach
requires a ``mean correction step'' to avoid numerical instability. In
practice, this entails periodically spreading information about each robot's
{estimate of} vector $\mathbf{x}$ entry across the team.

\subsection{Robustness Improvement Contribution}
\label{sec:robustness}

Motivation for adding a robustness contribution $u_i^r$ to control law
(\ref{eq:law}) was given in~\cite{ghedini2017b}.  A communication graph with a
positive $\lambda_2$ can be globally connected but still very susceptible to
the failures of nodes with high centrality scores (e.g., betweenness
centrality)~\cite{ghedini2017b}.  Robustness aims at mitigating this
vulnerability---critical for field experiments---quantified through the
heuristic {$\nu_i^k = \tfrac{|Path_{i}(k)|}{|\Pi_i|}$} where $|\Pi_i|$ is the number of
1- and 2-hops neighbors (see Figure~\ref{fig:theory}) of $i$, and $|Path_{i}(k)|$ is the number of nodes
that are exactly 2-hops away from node $i$ and relying on $\leq k$ 2-hops
paths to communicate with $i$.  Having defined {$q^k_i\in\mathbb{R}^3$} as the
barycentre of the robots in $Path_{i}(k)$, we compute the control contribution
as:
\begin{equation}
	u_i^r = \xi_r(\nu_i^k) \dfrac{q^k_i - p_i}{\left\| q^k_i - p_i \right\|}
	\label{eq:rob1}
\end{equation}
where $\xi_r(\cdot)$ evaluates as 0 or 1 depending on whether $V_i^k$
surpasses threshold $r$ or not~\cite{ghedini2017b}. The decentralized computation of $u_i^r$
requires the robots to know about their 2-hop neighbors, i.e., to be able to
exchange information about all their direct neighbors to all other members of
this same neighborhood.

\subsection{Coverage Improvement Contribution}
\label{sec:coverage}

The role of coverage contribution $u_i^{LJ}$ in \eqref{eq:law} is to
homogeneously spread robots over an area of interest 
as well as to provide simple collision avoidance by introducing repulsive forces between nearby robots that grow quickly as robots get closer.  The Lennard-Jones potential is a
simple, well-known inter-molecular interaction model whose control contribution
can be computed by deriving its expression and accounting for multiple
neighbors as follows:
\begin{equation}
	u^{LJ}_i = \sum_{n \in \mathcal{N}(i)} -\iota \left(  \left( \tfrac{a\cdot \delta^a}{\left(p_n-p_i\right)^{a+1}} \right)^{a} - 2 \cdot  \left( \tfrac{b \cdot \delta}{\left(p_n-p_i\right)^{b+1}} \right)^{b}  \right) 
	\label{eq:cov} 
\end{equation}
where $a$, $b$, $\delta$, and $\iota$ are the potential's parameters and
$\mathcal{N}(i)$ is the direct neighborhood of $i$. The decentralized
computation of $u_i^{LJ}$ only requires the 1-hop neighbors' positions---known under the situated communication model assumption.

\subsection{Simulations and Laboratory Experiments}
\label{sec:previous}

The control law in \eqref{eq:law} was originally implemented and
evaluated in a purely virtual environment: through MATLAB numerical
simulations in a 50x50 meter arena with 20 robots, a communication
range of 16 meters, and up to 70\% individual failures~\cite{ghedini2017b}.  A
first step towards a more realistic implementation was done
in~\cite{panerati2018}, using the multi-physics simulator ARGoS and 8 virtual
Footbots to optimize the hyper-parameters $\sigma$, $\psi$, and $\zeta$.
Also in~\cite{panerati2018}, \eqref{eq:law} was implemented in a team
of terrestrial robots (8 K-Team Khepera IV) and its performance evaluated, in
an uncluttered laboratory environment, against the injection of two types of
errors: (independent, exponentially distributed) robotic hardware failures and
packet drops (independent Bernoulli trials) in the communication links. We used a similar setup to investigate the distributed, online optimization of
the hyper-parameter~\cite{minelli2019,ras19si} and to improve robustness
through the coverage approach~\cite{siligardi2019}---switching from a
Lennard-Jones potential-based approach to a Voronoi tessellation.
Nonetheless, the major conceptual frailty of these experimental campaigns lays
in the fact that robot-to-robot communication was only emulated by
a central server.
In this work, we overcome this limitation by studying
an implementation %
that exploits actual point-to-point
communication based on DIGI's Xbee sub-1GHz RF
modules.

\section{Field Experiments}
\label{sec:field}

The disconnect between theoretical research and field robotics is often referred to as the
reality gap.  The field deployment and experiments described in what follows
are the major contributions of this work.  First, we developed the computing
hardware and software framework to support the control law presented in
Section~\ref{sec:law} in a team of quadcopters.  In particular, our software
implementation focuses on the message passing required by the decentralized
algorithms behind the three control contributions
\eqref{eq:conn}--\eqref{eq:cov}.  All necessary middleware---in the
form of ROS nodes to interface our software with the flight controller and the
XBee sub-1GHz RF modules---was also developed within Polytechnique Montreal's
MIST Laboratory.  Field experiments were conducted in Lanzarote, Spain during PANGAEA-X~\cite{ram}\footnote{\url{http://blogs.esa.int/caves/2018/12/04/a-swarm-of-drones/}}.

PANGAEA is the yearly geology training campaign organized by the European
Space Agency for its astronauts.  PANGAEA-X is an extension of this campaign
giving the opportunity to universities and researchers to deploy and test
their technologies in ``scenarios that mimic human and robotic operations away
from our planet''.  Because of its stringent fault-tolerance requirements and
communication delays, space exploration beyond low Earth orbit is one of the
applications that could benefit from decentralized multi-robot systems.

\subsection{Robotic and Computing Hardware}
\label{sec:hardware}

Our robotic platform is the Spiri, a small quadrotor
designed by Pleiades Robotics and intended for research and development.  The
Spiri is approximately 40$\times$40$\times$15 centimetres and weighs
1.5 kg.  Its flight controller is the PixRacer R14 which
interfaces to three additional modules: a compass and GPS/GLONASS receiver, a range finder
(to measure height) and a 2.4GHz RF module to interact with its remote
controller.
The companion on-board computer is an NVIDIA Jetson TX2 board with 8GB or LPDDR4 RAM, a hex-core ARMv8 CPU, and a 256-core Pascal GPU.
As an operating system (OS), we a use stripped-down version of the 64-bit release of Ubuntu 16.04.6 LTS Xenial Xerus, installed through NVIDIA's JetPack SDK.
A separate laptop, also running a Debian-based OS, acts as our ground station and interacts with the Spiris' Jetson TX2 boards through 5GHz 802.11n Wi-Fi (before flight) and a Digi XBee PRO900/SX868 sub-1GHz RF module (during flight).
The ground station initiates take-off and acts as a safeguard, offering backup control to the drone team.
These RF modules are also used on each Spiri for robot-to-robot communication.

\tikzset{
  box1/.style={
           rectangle,
           draw=black, 
           very thick,
           text centered},
    >=stealth',
    arrow1/.style={
           thick,
           shorten <=2pt,
           shorten >=2pt,
           }
}

\begin{figure}[!htb]
	\includegraphics[width=\columnwidth] {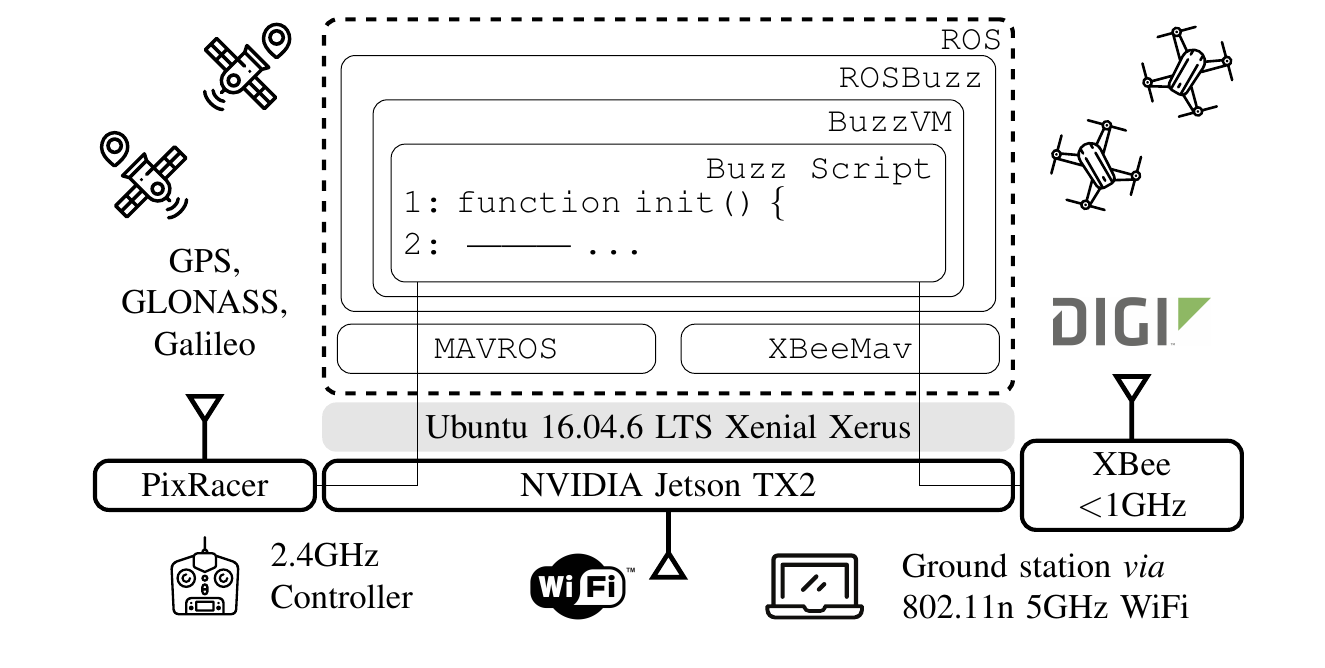}
\caption{
Block diagram summarizing the hardware components and software modules onboard each of the Pleiades Robotics' Spiri quadcopters (see Subsections~\ref{sec:hardware} and~\ref{sec:middleware}) 
}
\label{fig:diagram} 
\vspace{-1em}
\end{figure}

\subsection{Middleware and Software Implementation}
\label{sec:middleware}

For the software implementation of the decentralized control law in
Section~\ref{sec:law}---and the corresponding communication strategy described
below\footnote{\url{https://github.com/MISTLab/MessagePassing}}---we used the swarm-specific scripting language Buzz\footnote{\url{https://github.com/MISTLab/Buzz}} created by
Pinciroli and Beltrame~\cite{pinciroli2016}.  Buzz includes primitives
supporting the implementation of typical swarm robotics operations such as
polling from and broadcasting to all direct neighbors.  The language has a
simple syntax and was designed to allow researchers to create concise and
composable programs. These can be executed in teams of (possibly
heterogeneous) robots thanks to a portable, C-based virtual machine (VM).  The
VM allows to run
Buzz scripts
on multiple
platforms such as the Khepera IV, the Matrice 100, and the
Spiri.

The Jetson TX2 computers onboard each Spiri run ROS Kinetic Kame and the MAVROS node to needed communicate with the flight controller.
We then add two custom ROS nodes\footnote{\url{https://github.com/MISTLab/ROSBuzz}}\footnote{\url{https://github.com/MISTLab/XbeeMav}}: ROSBuzz~\cite{rosbuzzarxiv} and
XBeeMav.
The former is a node encapsulating the Buzz VM to interface it with the PixRacer flight controller and other ROS nodes. ROSBuzz also supports RVO collision avoidance.
XBeeMav is a node interfacing ROSBuzz with the XBee RF module
for serializing Buzz messages into MAVlink standard payloads.

Having this infrastructure in place, we want to study the feasibility of
implementing \eqref{eq:law} in a team of quadcopters.
In particular, we want to evaluate the performance of the information exchanges
needed for the decentralized computation of each one of the control contributions $u^c$, $u^r$, and $u^{LJ}$.

The connectivity improvement contribution $u^c$ (Subsection~\ref{sec:connectivity}) requires the estimation of $\lambda_2$.
Executing the decentralized PI update, as explained in~\cite{bertrand2013a}, needs a mean correction step.
To make this possible, all robots are required to re-broadcast information so that it can be spread over multiple communication hops.
In Buzz, this can be done with a \texttt{broadcast} call within a \texttt{listen} call.
This entails having information traveling possibly as many hops as the diameter of the communication graph. 
The mean correction step only needs to be performed periodically, for numerical stability.
The coverage control contribution $u^{LJ}$ (Subsection~\ref{sec:coverage}) is the simplest to compute as it only requires information about the positions of 1-hop neighbors.
This information in natively available within the runtime of Buzz (in a global \texttt{neighbor} structure).
In this case, messaging does not have to be dealt with explicitly because it is managed by the virtual machine.
Finally, the robustness improvement contribution $u^r$ (Subsection~\ref{sec:robustness}) is computed from the position information of 1- and 2-hop neighbors.
As Buzz makes 1-hop information readily available 
, to diffuse 2-hop information, robots only need to further broadcast it 
once 
and listen to the corresponding messages from direct neighbors.
\section{Results and Discussion}
\label{sec:results}

Our experiments were conducted using three Spiri quadcopters christened Mars, Pluto, and Valmiki.
The flight area was set on the island of Lanzarote approximately 5 kilometres north-east of PANGAEA's main site in a 300$\times$300 metres open field around coordinates 29.067$^\circ$N, 13.662$^\circ$W.
After two preliminary flights, all three drones were flown
for about 350 seconds (roughly 50\% of their ideal maximum flight time using 1600mAh battery packs) under manual control while, at the same time, running the infrastructure and Buzz implementation described in Section~\ref{sec:field}.
These experiments were meant to selectively stress-test the communication by forcing the drones to reach---large and small---inter-robot distances from which they would not have interacted, had they been solely controlled by \eqref{eq:law}.
The data collection process was aimed at 
verifying that our field setup could achieve the communication performance required to compute all three contributions of the law in \eqref{eq:law}.
Figure~\ref{fig:exp3-pos} presents the drones' trajectories, coordinates and inter-robot distances.
\begin{figure}[tb]%
	\centering
    	\includegraphics[] {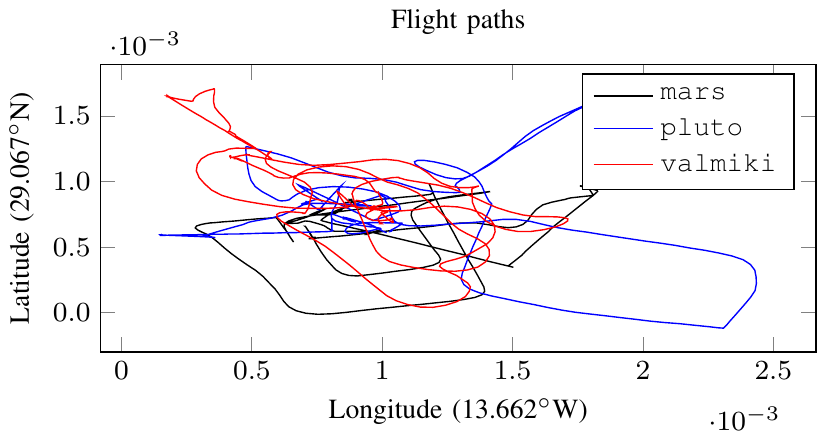}
    
    	\includegraphics[] {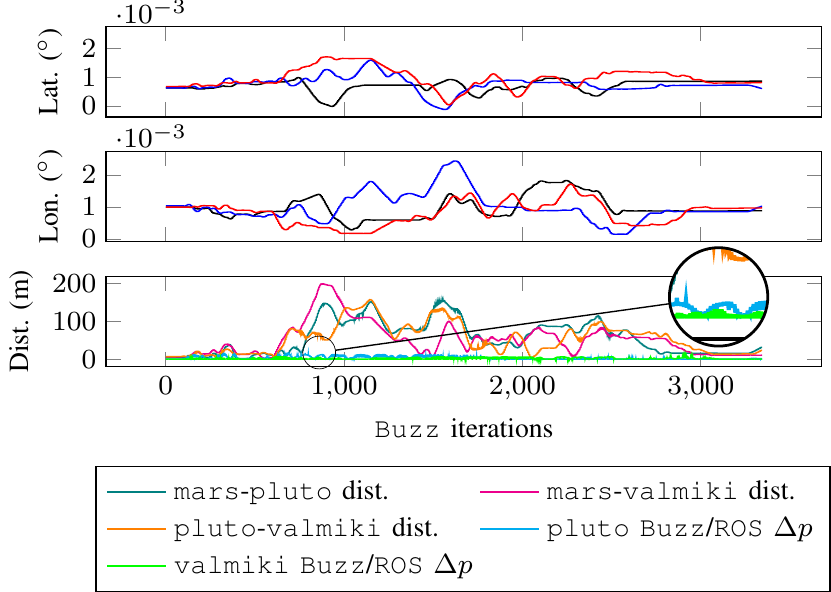}
	
\caption{
From top to bottom: (i) the quadcopters' trajectories; their (ii) latitude and (iii) longitude; (iv) the inter-robot distances and the discrepancies in position $\Delta p$ between the information stored in Buzz's logs and \texttt{rosbag} due to imperfect synchronization.
}
\label{fig:exp3-pos} 
\vspace{-1em}
\end{figure}

\subsection{Timing Performance}
\label{sec:timing}

One should observe that both Ubuntu and ROS are best-effort rather than real-time operating systems.
Hence, a first step in assessing the relevance of our experimental results required to verify the synchronization between by the operations of ROS, the Buzz VM, and the actual passing of time.
Figure~\ref{fig:exp3-synch} compares the evolution of the latitude and longitude logs---within Buzz, ROS, and with respect to the elapsed time---for two drones (Pluto and Valmiki).
We observe that Buzz deviates by 1\% or less from its ideal frequency of 10Hz. Thus, our implementation provides, if not real-time, at least timely execution. In the plots of this section, we use \texttt{Buzz} iterations as the abscissae.

\begin{figure}[!htb]%

	\includegraphics[] {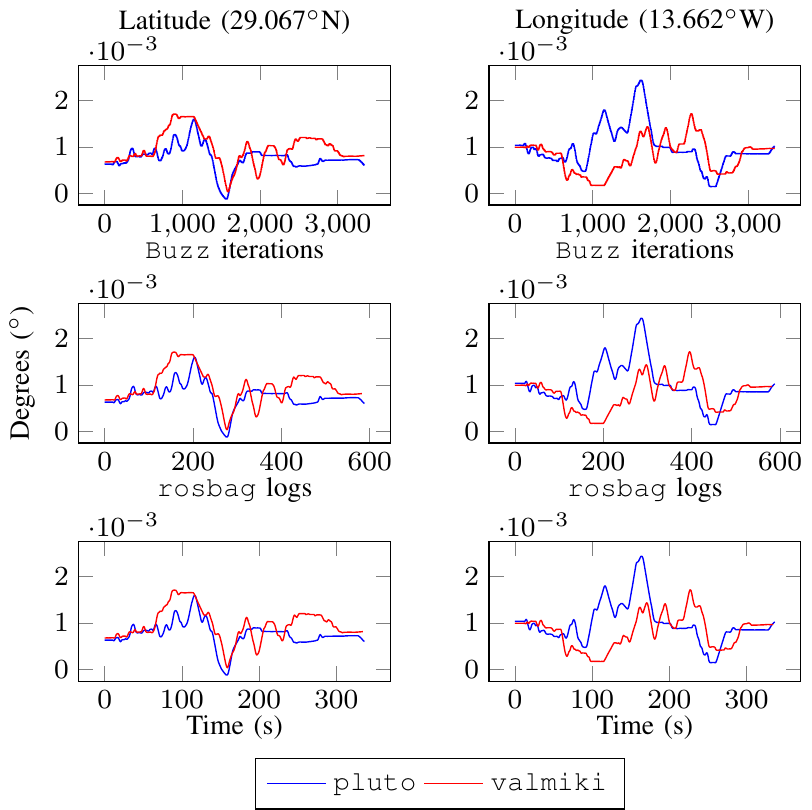}
	    
\caption{
Comparison of the evolution of latitude and longitude (from the experiment in Figure~\ref{fig:exp3-pos}) of Pluto and Valmiki against the progression of the Buzz VM, the \texttt{rosbag} log, and the absolute elapsed time when using a best-effort operating system.
}
\label{fig:exp3-synch} 
\vspace{-1em}
\end{figure}

\subsection{Connectivity}
\label{sec:connectivity-res}

Figure \ref{fig:exp3-alg} presents the results associated to the message passing required to compute $u^c$. 
The three charts in the left column present, for each one of the robots, the number of received messages originating from different robots per every line of a textual log (these logs have $\sim$5000 entries as they can be written more than once in a single Buzz iteration, if multiple messages were queued).
In an idealized, synchronous world, the number of such messages would steadily be 2. 
In practice, we observe that the plots constantly oscillate between 1 and 2. Yet, they are never 0, suggesting that the exchanges never broke down (at least, not until the end of the experiments, when robots were turned off).

\begin{figure}[tb]%
	   
	\includegraphics[] {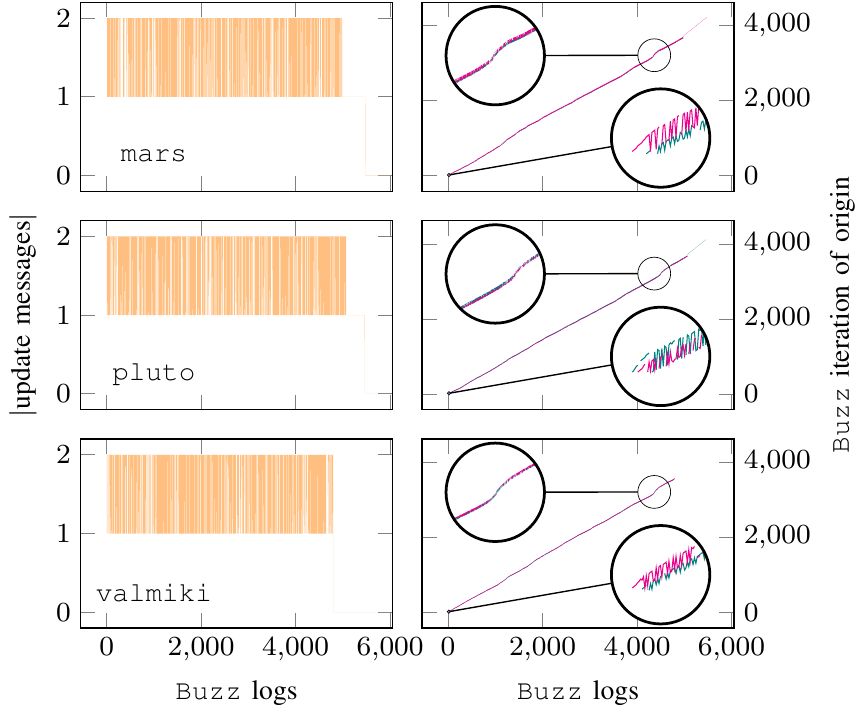}
    
\caption{
Performance results of the message passing required 
for the decentralized computation of the connectivity maintenance contribution $u^c$(Subsection~\ref{sec:connectivity}) of the control law in \eqref{eq:law}.
The left column shows the number of messages received by each robot while the right column displays their recentness (the magenta and teal lines representing the two different neighbors of origin).
}
\label{fig:exp3-alg} 
\vspace{-1em}
\end{figure}

\begin{table}%
\centering
\caption{
Buzz iterations (ratios) missing any of the 2-hop information messages. Correlations are computed until the 3000-th iteration, from the data in Figure~\ref{fig:exp3-rob}.
}

\begin{tabular}{ r  c  c  c  c }
\toprule

& \multicolumn{2}{p{2.5cm}}{\texttt{Buzz} iter. lacking 1 \texttt{robustness} mess. }  & \multirow{3}{*}{\parbox{0.8cm}{\centering \texttt{A}-\texttt{B} corr.}} & \multirow{3}{*}{\parbox{2.5cm}{\centering \texttt{Buzz} iter. without \texttt{robustness} mess.}} \\
\cmidrule(lr){2-3}
& From \texttt{A} & From \texttt{B} & & \\

\cmidrule(lr){2-2} \cmidrule(lr){3-3} \cmidrule(lr){4-4} \cmidrule(lr){5-5}

\texttt{mars} & $0.240$ & $0.266$ & $-0.115$ & $0.088$ \\
\texttt{pluto} & $0.265$ & $0.255$ & $+0.129$ & $0.051$  \\
\texttt{valmiki} & $0.236$ & $0.308$ & $+0.131$ & $0.052$ \\

\bottomrule
\end{tabular}
\label{tab:rob-gaps}
\end{table}
The charts in the right column of Figure \ref{fig:exp3-alg} present the evolution of the Buzz iteration of origin of each of these messages. For each robot, the two lines (teal and magenta) in the three plots refer to different senders (the two neighbors).
We can observe that, as time goes by, the received information stays current, i.e.,
originates in more recent Buzz iterations.
Once again, in an ideal world, these trends would be perfectly linear and monotone, with 
constant positive slopes.
In reality, we notice the presence of non-linear trends and very
small oscillations (whose detail is magnified) caused by the recursive way in which we relay messages, making it possible for slightly older information to bounce over multiple hops and to reach a robot after the most up-to-date one.
The overall trends indicate that the information needed for the mean correction of \eqref{eq:pi1} can be spread across the team but timing might become an issue for rapidly changing topologies.

\subsection{Robustness}
\label{sec:robustness-res}

The decentralized computation of the robustness improvement contribution $u^r$ in \eqref{eq:rob1} requires the relative positions of both 1- and 2-hop neighbors.
Sharing this information involves larger custom messages and the effectiveness
of the implementation required to compute $u^r$ is presented in Figure~\ref{fig:exp3-rob} for all three drones (the top six plots) versus the evolution the inter-robot distances (the bottom plot).
Table~\ref{tab:rob-gaps} summarizes, for each robot, the percentages of Buzz iterations in which either one or both messages coming from direct neighbors were not received, as well as the correlations between the lack of these message.

Similarly to the oscillations observed in Figure~\ref{fig:exp3-alg},
we can see in Figure~\ref{fig:exp3-rob} that, for all three robots, the number of direct neighbors
oscillates (between 1 and 2) and so does the number of indirect (2-hop) neighbors (between 2 and 4). 
Notably, more frequent drops in 1- and 2-hop neighbors in Figure~\ref{fig:exp3-rob} coincide
with periods of greater inter-robot distances and
the very end of our experiments, after the robots have landed.
This latter phenomenon  is likely explained by the joint negative effect of low battery and the ground obstructing the antennas.
The very low correlations between the lack of messages from 1-hop neighbors in Table~\ref{tab:rob-gaps} also suggest that these drops are more likely ascribed to external, independent causes (e.g., inter-robot distances) rather than intrinsic ones (e.g., a computational bottleneck). This indicates that computing $u^r$, can present scalability issues in larger robotic teams.

\begin{figure}[!htb]%
	    
	\includegraphics[] {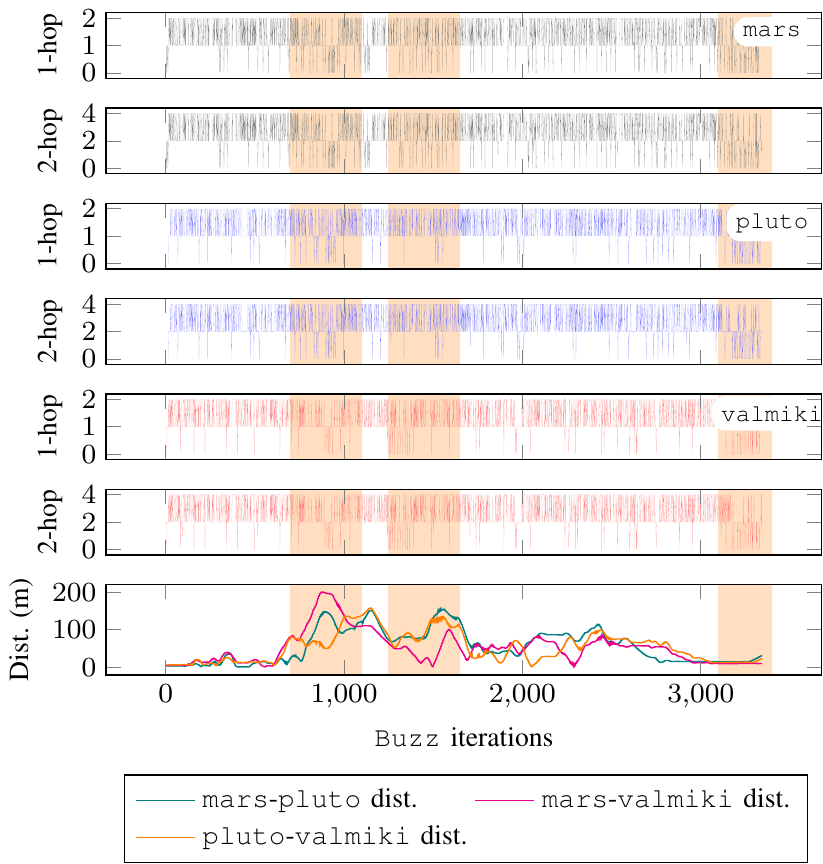}
	    
\caption{
Performance results of the message passing required 
for the decentralized computation of the robustness improvement contribution $u^r$(Subsection~\ref{sec:robustness}).
The number of 1- and 2-hop neighbors (including themselves) known to each robot are plotted against the inter-robot distances.
}
\label{fig:exp3-rob} 
\vspace{-1em}
\end{figure}

\subsection{Coverage}
\label{sec:coverage-res}

As we explained in Subsection \ref{sec:middleware}, the coverage improvement
contribution $u^{LJ}$ in \eqref{eq:cov} is the simplest to compute
in a decentralized fashion as it only requires information about the relative
positions of all direct neighbors of a drone.  Figure~\ref{fig:exp3-lej} shows
how this information evolves over time on-board each robot.  We do so by
plotting each robot's on-board, presumed inter-robot distances against the GPS-given ground truth---the bottom
chart.  We observe an almost perfect match: the robots only
sporadically lose track of their neighbors for fractions of
seconds (the zoomed-in bubbles), meaning that they can reliably compute $u^{LJ}$.

\begin{figure}[htb]%
	    
	\includegraphics[] {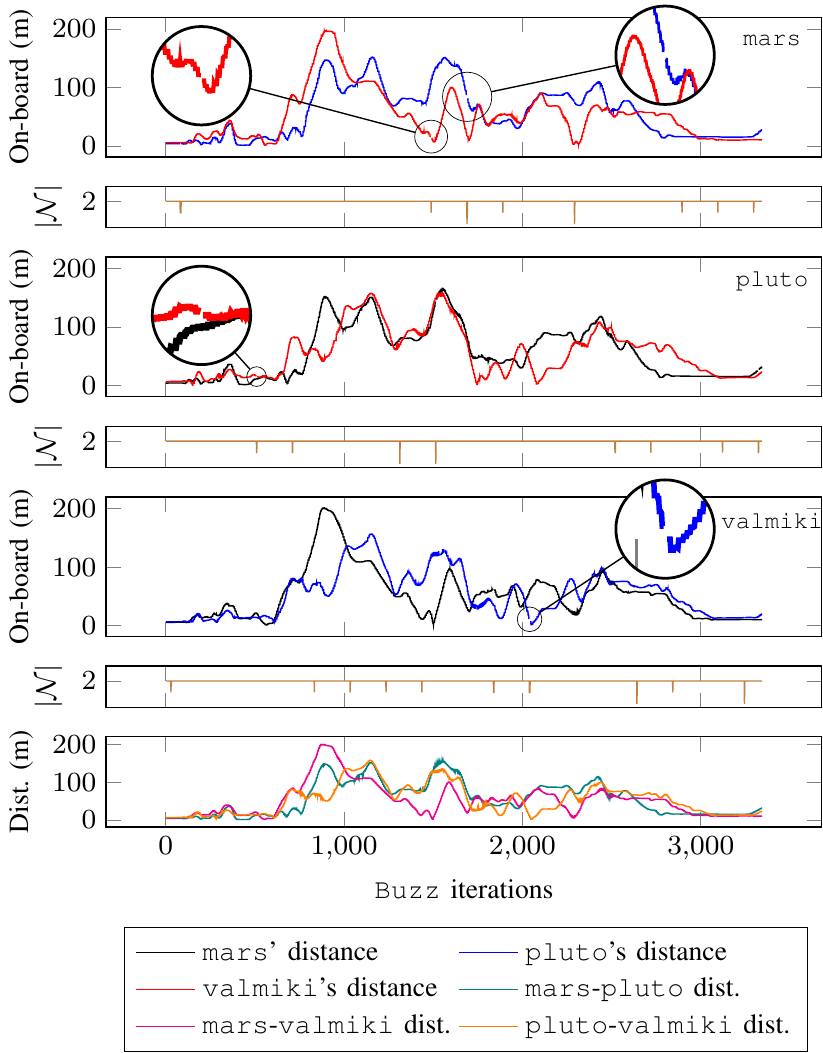}
    
\caption{
Performance results of the message passing required 
for the decentralized computation of the coverage improvement contribution $u^{LJ}$(Subsection~\ref{sec:coverage}) of the control law in \eqref{eq:law}.
The estimated inter-robot distances onboard each robot are compared with the ground truth (the bottom plot). The brown lines show the number of entries stored within Buzz's \texttt{neighbor} structure.
}
\label{fig:exp3-lej} 
\vspace{-1em}
\end{figure}

\section{Conclusions}
\label{sec:conclusions}

In this paper, 
we tackled the %
reality gaps associated to decentralized, robust, global connectivity control laws in a multi-robot system using three quadcopters communicating with sub-1GHz RF modules.
Prior to this work, most of the research in the area had only focused on numerical simulations and indoor experiments.
Our first contribution was the creation of the hardware and software stack implementing the control law proposed in~\cite{ghedini2017b}.
Then, we brought this stack to a team of quadcopters and performed field tests (in the context of ESA's PANAGEA-X training campaign) to assess the performance of our implementation, especially with respect to information exchanges.
Our results are encouraging as they indicate that the information required to compute all three components of the decentralized control law in Equation~\ref{eq:law} can be spread across multiple robots even when flying hundreds of meters apart.
Yet, these tests also show that the reality gap---with respect to assumptions on communication made by previous simulation~\cite{ghedini2017b} and laboratory~\cite{panerati2018} studies---is remarkable as, oftentimes, only part of the total information is available to each robot.

\balance

\end{document}